\documentclass[11pt,a4paper]{article}
\usepackage[hyperref]{acl2021}
\usepackage{times}
\usepackage{latexsym}

\usepackage{tabularx}
\makeatletter
\def\hlinewd#1{%
\noalign{\ifnum0=`}\fi\hrule \@height #1 %
\futurelet\reserved@a\@xhline}
\makeatother

\usepackage{microtype}

\aclfinalcopy 

\usepackage{adjustbox}
\usepackage{multicol}  
\usepackage{multirow} 
\usepackage{amsmath}
\usepackage{amsfonts}
\usepackage{makecell}

\usepackage{verbatim}
\usepackage{mathtools}

\usepackage{booktabs}

\usepackage{booktabs}
\usepackage{makecell}
\usepackage{ulem}
\usepackage{cleveref}
\usepackage{ulem}

\usepackage{times}
\usepackage{latexsym}

\usepackage{algpseudocode} 

\usepackage[hang,flushmargin]{footmisc}

\usepackage[linesnumbered,ruled]{algorithm2e}
\SetAlFnt{\small}
\SetAlCapNameFnt{\normalsize}
\makeatletter
\newcommand{\nosemic}{\renewcommand{\@endalgocfline}{\relax}}
\newcommand{\dosemic}{\renewcommand{\@endalgocfline}{\algocf@endline}}
\let\oldnl\nl
\newcommand{\nonl}{\renewcommand{\nl}{\let\nl\oldnl}}
\makeatother

\SetAlCapFnt{\small}
\SetAlCapNameFnt{\small}

\usepackage{tabularx}
\makeatletter
\def\hlinewd#1{%
\noalign{\ifnum0=`}\fi\hrule \@height #1 %
\futurelet\reserved@a\@xhline}
\makeatother

\usepackage{algpseudocode}  
\usepackage{amsmath}
\usepackage{multicol}  
\usepackage{multirow} 
\usepackage{flexisym}
\usepackage{graphicx}  
\usepackage{array}
\newcolumntype{L}[1]{>{\raggedright\let\newline\\\arraybackslash\hspace{0pt}}m{#1}}
\newcolumntype{C}[1]{>{\centering\let\newline\\\arraybackslash}m{#1}}
\newcolumntype{R}[1]{>{\raggedleft\let\newline\\\arraybackslash\hspace{0pt}}m{#1}}

\usepackage{cleveref}
\crefformat{section}{\S#2#1#3} 
\crefformat{subsection}{\S#2#1#3}
\crefformat{subsubsection}{\S#2#1#3}

\title{Dialogue Response Selection with Hierarchical Curriculum Learning}
\author{Yixuan Su$^{\spadesuit,}$\thanks{~~The main body of this work was done during internship at Tencent Inc. The first two authors contributed equally. Yan Wang is the corresponding
author.}~\quad Deng Cai$^{\heartsuit,*}$\quad  Qingyu Zhou$^{\diamondsuit}$\quad Zibo Lin$^{\diamondsuit}$\quad Simon Baker$^{\spadesuit}$\quad \\
\textbf{Yunbo Cao}$^{\diamondsuit}$\quad \textbf{Shuming Shi}$^\diamondsuit$\quad \textbf{Nigel Collier}$^\spadesuit$\quad \textbf{Yan Wang}$^{\diamondsuit}$  \\
$^\spadesuit$Language Technology Lab, University of Cambridge \\
$^\heartsuit$The Chinese University of Hong Kong \\
$^\diamondsuit$Tencent Inc.\\

\\ {\tt \{ys484,sb895,nhc30\}@cam.ac.uk}\\ {\tt thisisjcykcd@gmail.com} \\ {\tt \{qingyuzhou,jimblin,yunbocao,shumingshi,brandenwang\}@tencent.com}\\
}

\date{}

\begin{document}
\maketitle
\begin{abstract}
We study the learning of a matching model for dialogue response selection. Motivated by the recent finding that models trained with random negative samples are not ideal in real-world scenarios, we propose a hierarchical curriculum learning framework that trains the matching model in an ``easy-to-difficult" scheme. Our learning framework consists of two complementary curricula: (1) corpus-level curriculum (CC); and (2) instance-level curriculum (IC). In CC, the model gradually increases its ability in finding the matching clues between the dialogue context and a response candidate. As for IC, it progressively strengthens the model's ability in identifying the mismatching information between the dialogue context and a response candidate. Empirical studies on three benchmark datasets with three state-of-the-art matching models demonstrate that the proposed learning framework significantly improves the model performance across various evaluation metrics.
\end{abstract}

\section{Introduction}
\label{sec:intro}
Building intelligent conversation systems is a long-standing goal of artificial intelligence and has attracted much attention in recent years \cite{DBLP:journals/corr/abs-1801-01957,DBLP:conf/naacl/KollarBSOCMKSM18}. An important challenge for building such conversation systems is the response selection problem, that is, selecting the best response to a given dialogue context from a set of candidate responses \cite{DBLP:conf/emnlp/RitterCD11}.

\begin{table}[tb]
\begin{center}
\resizebox{0.98\columnwidth}{!}{
\begin{tabular}{|l|}
\hline
\textbf{Dialogue Context Between Two Speakers A and B} \\
\hline
\textbf{A}: Would you please recommend me a good TV series \\\hspace{1em} to watch during my spare time?  \\

\textbf{B}: Absolutely! Which kind of TV series are you most\\\hspace{1em} interested in?  \\
\textbf{A}: My favorite type is fantasy drama. \\
\textbf{B}: I think both Game of Thrones and The Vampire \\\hspace{1em} Diaries are good choices.  \\
\hline

\makecell[c]{\textbf{Positive Response}}  \\
\hline
\textbf{P1}: Awesome, I believe both of them are great TV \\\hspace{1em} \; series! I will first watch Game of Thrones. (\textbf{Easy})\\
\textbf{P2}: Cool! I think I find the perfect thing to kill my\\\hspace{1em} \; weekends. \; \; \; \; \; \; \; \; \; \; \; \; \; \; \; \; \; \; \; \;(\textbf{Difficult})\\
\hline
\makecell[c]{\textbf{Negative Response}}  \\
\hline
\textbf{N1}: This restaurant is very expensive. \; \; \; \; \; \; (\textbf{Easy})\\
\textbf{N2}: Iain Glen played Ser Jorah Mormont in the HBO \\\hspace{1em} \; fantasy series Game of Thrones. \; \; \; \; \;(\textbf{Difficult})\\
\hline
\end{tabular}
}
\end{center}
  \caption{An example dialogue context between speakers A and B, where P1 and P2 are easy and difficult positives; N1 and N2 are easy and difficult negatives.}
\label{tb:retrieved_example}
\end{table}

To tackle this problem, different matching models are developed to measure the matching degree between a dialogue
context and a response candidate \cite{DBLP:conf/acl/WuWXZL17,DBLP:conf/acl/WuLCZDYZL18,DBLP:conf/acl/LuZXLZX19,DBLP:conf/cikm/GuLL19}. Despite their differences, most prior works train the model with data constructed by a simple heuristic. For each context, the human-written response is considered as positive (i.e., an appropriate response) and the responses from other dialogue contexts are considered as negatives (i.e., inappropriate responses). In practice, the negative responses are often randomly sampled and the training objective ensures that the positive response scores are higher than the negative ones.

Recently, some researchers \cite{li-etal-2019-sampling,lin2020world} have raised the concern that randomly sampled negative responses are often too trivial (i.e., totally irrelevant to the dialogue context). Models trained with trivial negative responses may fail to handle strong distractors in real-world scenarios. Essentially, the problem stems from the ignorance of the diversity in context-response matching degree. In other words, all random responses are treated as equally negative regardless of their different distracting strengths. For example, Table \ref{tb:retrieved_example} shows a conversation between two speakers and two negative responses (N1, N2) are presented. For N1, one can easily dispel its appropriateness as it unnaturally diverges from the TV show topic. On the other hand, N2 is a strong distractor as it overlaps significantly with the context (e.g., \textit{fantasy series} and \textit{Game of Thrones}). Only with close observation we can find that N2 does not maintain the coherence of the discussion, i.e., it starts a parallel discussion about an actor in \textit{Game of Thrones} rather than elaborating on the enjoyable properties of the TV series. In addition, we also observe a similar phenomenon on the positive side. For different training context-response pairs, their pairwise relevance also varies. In Table \ref{tb:retrieved_example}, two positive responses (P1, P2) are provided for the given context. For P1, one can easily confirm its validity as it naturally replies the context. As for P2, while it expatiates on the enjoyable properties of the TV series, it does not exhibit any obvious matching clues (e.g., lexical overlap with the context). Therefore, to correctly identify P2, its relationship with the context must be carefully reasoned by the model. \\\indent Inspired by the above observations, in this work, we propose to employ the idea of curriculum learning (CL) \cite{DBLP:conf/icml/BengioLCW09}. The key to applying CL is to specify a proper learning scheme under which all training examples are learned. By analyzing the characteristics of the concerned task, we tailor-design a hierarchical curriculum learning (HCL) framework. Specifically, our learning framework consists of two complementary curriculum strategies, corpus-level curriculum (CC) and instance-level curriculum (IC), covering the two distinct aspects of response selection. In CC, the model gradually increases its ability in finding matching clues through an easy-to-difficult arrangement of positive context-response pairs. In IC, we sort all negative responses according to their distracting strength such that the model’s capability of identifying the mismatching information can be progressively strengthened. \\\indent Notably, our learning framework is independent to the choice of matching models. For a comprehensive evaluation, we evaluate our approach on three representative matching models, including the current state of the art.
Results on three benchmark datasets demonstrate that the proposed learning framework leads to remarkable performance improvements across all evaluation metrics. \\\indent
In a nutshell, our contributions can be summarized as:
(1) We propose a hierarchical curriculum learning framework to tackle the task of dialogue response selection; and (2) Empirical results on three benchmark datasets show that our approach can significantly improve the performance of various strong matching models, including the current state of the art.

\begin{figure*}[t] 
	\centering    
	\setlength{\abovecaptionskip}{3pt}
\includegraphics[width=1.0\textwidth]{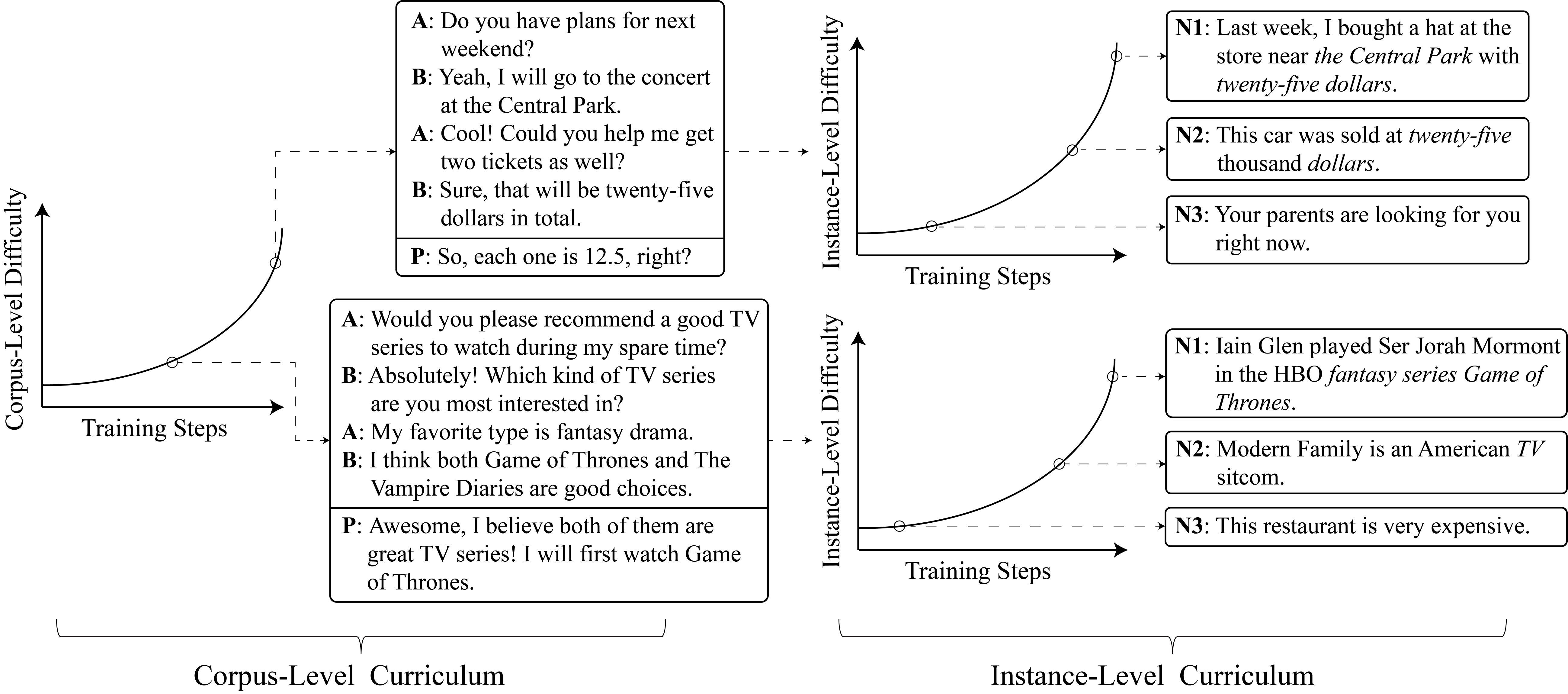}
	\caption{An illustration of the proposed learning framework: On the left part, two training context-response pairs with different difficulty levels are presented (the upper one is more difficult than the lower one, and \textbf{P} denotes the positive response). For each training instance, we show three associated negative responses (\textbf{N1}, \textbf{N2} and \textbf{N3}) with increasing difficulty from the bottom to the top. In the negative responses, the words that also appear in the dialogue context are marked as \textit{italic}.}
    \label{fig:overview_diagram}
\end{figure*}

\section{Background}
Given a dataset $\mathcal{D}=\{(c_i, r_i)\}_{i=1}^{|\mathcal{D}|}$, the learning of a matching model $s(\cdot, \cdot)$ is to correctly identify the positive response $r_i$ conditioned on the dialogue context $c_i$ from a set of negative responses $\mathcal{R}_i^-$. The learning objective is typically defined as 
\begin{equation}
\label{eq:matching_model}
    \mathcal{L}_s = \sum_{j=1}^m\max\{0,1 - s(c_i, r_i) + s(c_i, \mathcal{R}_{ij}^-)\},
\end{equation}
where $m$ is the number of negative responses associated with each training context-response pair. In most existing studies  \cite{DBLP:conf/acl/WuWXZL17,DBLP:conf/acl/WuLCZDYZL18,DBLP:conf/cikm/GuLL19}, the training negative responses $\mathcal{R}_i^-$ are randomly selected from the dataset $\mathcal{D}$. Recently, \citet{li-etal-2019-sampling} and \citet{lin2020world} proposed different approaches to strengthen the training negatives. In testing, for any context-response $(c,r)$, the models give a score $s(c,r)$ that reflects their pairwise matching degree. Therefore, it allows the user to rank a set of response candidates according to the scores for response selection.

\section{Methodology}
\subsection{Overview}
We propose a \textit{hierarchical curriculum learning} (HCL) framework for training neural matching models. It consists of two complementary curricula: (1) corpus-level curriculum (CC); and (2) instance-level curriculum (IC). Figure \ref{fig:overview_diagram} illustrates the relationship between these two strategies. In CC (\cref{sec:cc}), the training context-response pairs with lower difficulty are presented to the model before harder pairs. 
This way, the model gradually increases its ability to find the matching clues contained in the response candidate.
As for IC (\cref{sec:ic}), it controls the difficulty
of negative responses that associated with each training context-response pair. Starting from easier negatives, the model progressively strengthens its ability to identify the mismatching information (e.g., semantic incoherence) in the response candidate. 
The following gives a detailed description of the proposed approach.

\subsection{Corpus-Level Curriculum}
\label{sec:cc}
Given the dataset $\mathcal{D}=\{(c_i, r_i)\}_{i=1}^{|\mathcal{D}|}$, the corpus-level curriculum (CC) arranges the ordering of different training context-response pairs. The model first learns to find easier matching clues from the pairs with lower difficulty. As the training progresses, harder cases are presented to the model to learn less obvious matching signals. Two examples are shown in the left part of Figure \ref{fig:overview_diagram}. For the easier pair, the context and the positive response are semantically coherent as well as lexically overlapped (e.g., \textit{TV series} and \textit{Game of Thrones}) with each other and such matching clues are simple for the model to learn. As for the harder case, the positive response can only be identified via numerical reasoning, which makes it harder to learn.

\begin{figure*}[t] 
	\centering    
	\setlength{\abovecaptionskip}{3pt}
\includegraphics[width=1.0\textwidth]{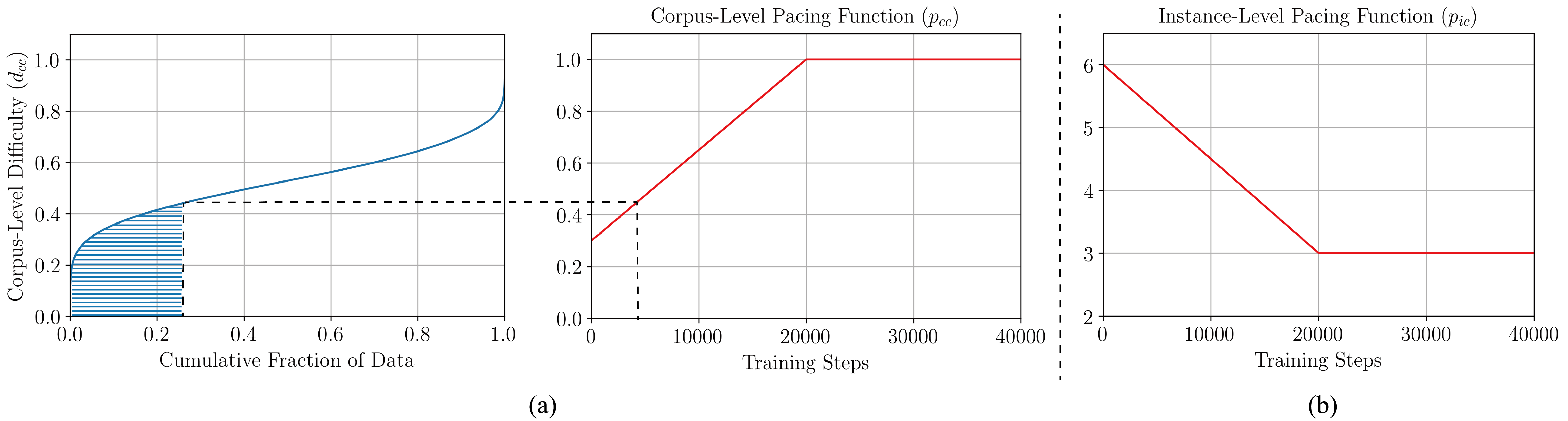}
	\caption{(a) Illustration of the corpus-level curriculum. At each step: (1) $p_{cc}(t)$ is computed based on the current step $t$; and (2) a batch of context-response pairs are uniformly sampled from the training instances whose corpus-level difficulty is lower than $p_{cc}(t)$ (shaded area in the example). In this example, $p_{cc}(0)=0.3$ and $T=20000$; (b) Illustration of the instance-level pacing function. In this example, $k_0=\log_{10}^{|\mathcal{D}|}=6$, $k_T=3$, and $T = 20000$.}
    \label{fig:cc}
\end{figure*}

\paragraph{Difficulty Function.} To measure the difficulty of each training context-response pair $(c_i, r_i)$, we adopt a pre-trained ranking model $G(\cdot, \cdot)$ (\cref{sec:ranker}) to calculate its relevance score as $G(c_i, r_i)$. Here, a higher score of $G(c_i, r_i)$ corresponds to a higher relevance between $c_i$ and $r_i$ and vice versa. Then, for each pair $(c_i, r_i)\in\mathcal{D}$, its corpus-level difficulty $d_{cc}(c_i, r_i)$ is defined as 

\begin{equation}
\label{eq:corpus-difficulty}
        d_{cc}(c_i, r_i) = 1.0 - \frac{G(c_i, r_i)}{\max_{(c_k, r_k)\in\mathcal{D}}G(c_k, r_k)},\\
\end{equation}
where $d_{cc}(c_i, r_i)$ is normalized to $[0.0,1.0]$. Here, a lower difficulty score indicates the pair $(c_i, r_i)$ is easier for the model to learn and vise versa.

\paragraph{Pacing Function.} In training, to select the training context-response pairs with appropriate difficulty, we define a corpus-level pacing function, $p_{cc}(t)$, which controls the pace of learning from easy to hard instances. In other words, at time step $t$, $p_{cc}(t)$ represents the upper limit of difficulty and the model is only allowed to use the training instances $(c_i, r_i)$ whose corpus-level difficulty score $d_{cc}(c_i, r_i)$ is lower than $p_{cc}(t)$. In this work, we propose a simple functional form for $p_{cc}(t)$\footnote{More sophisticated designs for the function $p_{cc}(t)$ are possible, but we do not consider them in this work.} as
$$ 
    p_{cc}(t) =
      \begin{cases} 
      \frac{1.0-p_{cc}(0)}{T}\cdot t + p_{cc}(0)& \mbox{if } t\leq T,\\
      1.0 & \textup{otherwise},
\end{cases} $$
where $p_{cc}(0)$ is a predefined initial value. At the training warm up stage (first $T$ steps), we learn a basic matching model with a easy subset of the training data. In this subset, the difficulty of all samples are lower than $p_{cc}(t)$. After $p_{cc}(t)$ becomes $1.0$ (at time step $T$), the corpus-level curriculum is completed and the model can then freely access the entire dataset. In Figure \ref{fig:cc}(a), we give an illustration of the corpus-level curriculum.

\subsection{Instance-Level Curriculum}
\label{sec:ic}
As a complement of CC, the instance-level curriculum (IC) controls the difficulty of negative responses. For an arbitrary training context-response pair $(c_i, r_i)$, while its associated negative responses can be any responses $r_{j}$ ($\textup{s.t. } j\ne i$) in the training set, the difficulties of different $r_j$ are diverse. Some examples are presented in the right part of Figure \ref{fig:overview_diagram}. We see that the negative responses with lower difficulty are always simple to spot as they are often obviously off the topic. As for the harder negatives, the model need to identify the fine-grained semantic incoherence between them and the context.

The main purpose of IC is to select negative responses with appropriate difficulty based on the state of the learning process. At the beginning, the negative responses are randomly sampled from the entire training set, so that most of them are easy to distinguish. As the training evolves, IC gradually increases the difficulty of negative responses by sampling them from the responses with higher difficulty (i.e., from a harder subset of the training data). In this way, the model's ability in finding the mismatching information is progressively strengthened and will be more robust when handling those strong distractors in real-world scenarios.

\paragraph{Difficulty Function.} Given a specific training instance $(c_i, r_i)$, we define the difficulty of an arbitrary response $r_{j}$ ($\textup{s.t. } j\ne i$) as its rank in a sorted list of relevance score in descending order, 
\begin{equation}
\label{eq:instance-difficulty}
    d_{ic}(c_i, r_j) = \textup{sort}_{r_j \in \mathcal{D}, j\ne i}(G(c_i, r_j)).
\end{equation}
In this formula, the response $r_h$ with the highest relevance score, i.e., $r_h=\max_{r_j\in\mathcal{D},j\ne i}G(c_i,r_j)$, has a rank of 1, thus $d_{ic}(c_i,r_h)=1$. For the response $r_l$ with the lowest relevance score, i.e., $r_l =\min_{r_j\in\mathcal{D},j\ne i}G(c_i,r_j)$, has a rank of $|\mathcal{D}|$, thus $d_{ic}(c_i,r_l)=|\mathcal{D}|$. Here, a smaller rank means the corresponding negative response is more relevant to the context $c_i$, thus it is more difficult for the model to distinguish.

\paragraph{Pacing Function.} 
Similar to CC, in IC, the pace of learning from easy to difficult negative responses is controlled by an instance-level pacing function, $p_{ic}(t)$. It adjusts the size of the sampling space (in log scale) 
from which the negative responses are sampled from. 
Given a training instance $(c_i, r_i)$, at time step $t$, the negative examples are sampled from the responses $r_{j}$ ($\textup{s.t. } j\ne i$) whose rank is smaller than $10^{p_{ic}(t)}$ ($d_{ic}(c_i, r_j)\leq 10^{p_{ic}(t)}$), i.e., the negative responses are sampled from a subset of the training data which consists of the top-$10^{p_{ic}(t)}$ relevant responses in relation to $c_i$. The smaller the $p_{ic}(t)$ is, the harder the sampled negatives will be. In this work, we define the function $p_{ic}(t)$ as 
$$ 
    p_{ic}(t) =
      \begin{cases} 
      \frac{(k_0 - k_T)}{T}\cdot(T-t) + k_T & \mbox{if } t\leq T,\\
      k_T & \mbox{if } t>T,
\end{cases} $$
where $T$ is the same as the one in the corpus-level pacing function $p_{cc}(t)$. $k_0=\log_{10}^{|\mathcal{D}|}$, meaning that, at the start of training, the negative responses are sampled from the entire training set $\mathcal{D}$. $k_T$ is a hyperparameter and it is smaller than $k_0$. After $p_{ic}(t)$ becomes $k_T$ (at step $T$), the instance-level curriculum is completed. For the following training steps, the size of the sampling space is fixed at $10^{k_T}$. An example of $p_{ic}(t)$ is depicted in Figure \ref{fig:cc}(b).

\normalem
\begin{algorithm}[t]
    \caption{Hierarchical Curriculum Learning}
    \SetKwInOut{Input}{Input}
    \SetKwInOut{Output}{Output}
    
    \Input{Dataset, $\mathcal{D}=\{(c_i, r_i)\}_{i=1}^{|\mathcal{D}|}$; model trainer, $\mathcal{T}$, that takes batches of training data as input to optimize the matching model; corpus-level difficulty and pacing function, $d_{cc}$ and $p_{cc}$; instance-level difficulty and pacing function, $d_{ic}$ and $p_{ic}$; number of negative responses, $m$;}

    \For{train step $t = 1, ...$}
      {
        \nosemic Uniformly sample one batch of context-response pairs, $B_t$, from all $(c_i, r_i)\in\mathcal{D}$, such that $d_{cc}(c_i, r_i)\leq p_{cc}(t)$, as shown in Figure \ref{fig:cc}(a).\;
        \For{$(c_j, r_j)$ in $B_t$}
        {
            \nosemic Sample $m$ negative responses, $\mathcal{R}_j^-$, from all responses $r$, where $r\ne r_j$, that satisfies the condition $d_{ic}(c_j, r)\leq 10^{p_{ic}(t)}$. 
        }
        Invoke the trainer, $\mathcal{T}$, using $\{(c_k, r_k, \mathcal{R}_k^-)\}_{k=1}^{|B_t|}$ as input to optimize the model using Eq. \eqref{eq:matching_model}.
      }
    \Output{Trained Matching Model}
\end{algorithm}

\subsection{Hierarchical Curriculum Learning}
\label{sec:ranker}

\paragraph{Model Training.} Our learning framework jointly employs the corpus-level and instance-level curriculum.
For each training step, we construct a batch of training data as follows: First, we select the positive context-response pairs according to the corpus-level pacing function $p_{cc}(t)$. Then, for each instance in the selected batch, we sample its associated negative examples according to the instance-level pacing function $p_{ic}(t)$. Details of our learning framework are presented in Algorithm 1.

\paragraph{Fast Ranking Model.} As described in Eq. \eqref{eq:corpus-difficulty} and \eqref{eq:instance-difficulty}, our framework requires a ranking model $G(\cdot,\cdot)$ that efficiently measures the pairwise relevance of millions of possible context-response combinations. In this work, we construct $G(\cdot,\cdot)$ as
an non-interaction matching model with dual-encoder structure such that we can precompute all contexts and responses offline and store them in cache. For any context-response pair $(c, r)$, 
its pairwise relevance $G(c, r)$ is defined as 
\begin{equation}
\label{eq:similarity}
    G(c, r) = E_c(c)^T E_r(r),
\end{equation}
where $E_c(c)$ and $E_r(r)$ are the dense context and response representations produced by a context encoder $E_c(\cdot)$ and a response encoder $E_r(\cdot)$\footnote{The encoders can be any model, e.g., LSTM \cite{DBLP:journals/neco/HochreiterS97} and Transformers \cite{DBLP:conf/nips/VaswaniSPUJGKP17}.}. 

\paragraph{Offline Index.} 
After training the ranking model on the same response selection dataset $\mathcal{D}$ using the in-batch negative objective \cite{DBLP:conf/emnlp/KarpukhinOMLWEC20}, we compute the dense representations of all contexts and responses contained in $\mathcal{D}$. Then, as described in Eq. \eqref{eq:similarity}, the relevance scores of all possible combinations of the contexts and responses in $\mathcal{D}$ can be easily computed through the dot product between their representations. After this step, we can compute the corpus-level and instance-level difficulty of all possible combinations and cache them in memory for a fast access in training.

\section{Related Work}
\label{sec:related}
\paragraph{Dialogue Response Selection.} Early studies in this area devoted to the response selection for single-turn conversations \cite{DBLP:conf/emnlp/WangLLC13,tan2016lstmbased,DBLP:journals/corr/abs-2004-02214}. Recently, researchers turned to the scenario of multi-turn conversations and many sophisticated neural network architectures have been devised \cite{DBLP:conf/acl/WuWXZL17,DBLP:conf/cikm/GuLL19,DBLP:conf/acl/WuLCZDYZL18,DBLP:conf/cikm/GuLLLSWZ20}. 

There is an emerging line of research studying how to improve existing matching models with better learning algorithms. \citet{DBLP:conf/acl/WuwLZ18} proposed to adopt a Seq2seq model as weak teacher to guide the training process. \citet{DBLP:conf/acl/FengTWFZY19} designed a co-teaching framework to eliminate the training noises. Similar to our work, \citet{li-etal-2019-sampling} proposed to alleviate the problem of trivial negatives by sampling stronger negatives. \citet{lin2020world} attempted to create  negative examples with a retrieval system and a pre-trained generation model. In contrast to their studies, we not only enlarge the set of negative examples but also arrange the negative examples in an easy-to-diffuclt fashion.

\begin{table*}[tb]
	\small
	\centering
	\renewcommand{\arraystretch}{1.15}
	\scalebox{0.805}{
	\begin{tabular}{lccccccccccccc}
		\hlinewd{0.75pt}
		\multicolumn{1}{c}{\multirow{2}{*}{\textbf{Model}}}    & \multicolumn{6}{c}{\textbf{Douban}}& \multicolumn{4}{c}{\textbf{Ubuntu}}&\multicolumn{3}{c}{\textbf{E-Commerce}}  \\ 
		\cmidrule(lr){2-7}
		\cmidrule(lr){8-11}
		\cmidrule(lr){12-14}
		& $\rm {MAP}$ & $\rm MRR$ & $\rm P$@$1$ & $\rm R_{10}$@$1$ & $\rm R_{10}$@$2$ & $\rm R_{10}$@$5$&$\rm R_{2}$@$1$&$\rm R_{10}$@$1$ & $\rm R_{10}$@$2$ & $\rm R_{10}$@$5$&$\rm R_{10}$@$1$ & $\rm R_{10}$@$2$ & $\rm R_{10}$@$5$  \\
		\hline
		RNN  &0.390 &0.422 &0.208 &0.118 &0.223 &0.589  &0.768 &0.403 &0.547 &0.819&0.325&0.463&0.775\\ 
		CNN &0.417 &0.440 &0.226 &0.121 &0.252 &0.647  &0.848 &0.549 &0.684 &0.896&0.328&0.515&0.792  \\
		LSTM& 0.485&0.527 &0.320 &0.187 &0.343 &0.720  &0.901 &0.638 &0.784 &0.949&0.365&0.536&0.828  \\
	    BiLSTM&0.479 &0.514 &0.313 &0.184 &0.330 &0.716  &0.895 &0.630 &0.780 &0.944&0.355&0.525&0.825  \\
		MV-LSTM&0.498 &0.538 &0.348 &0.202 &0.351 &0.710  &0.906 &0.653 &0.804 &0.946&0.412&0.591&0.857  \\
		Match-LSTM&0.500 &0.537 &0.345 &0.202 &0.348 &0.720  &0.904 &0.653 &0.799 &0.944&0.410&0.590&0.858  \\
		\hline
		DL2R& 0.488 &0.527 &0.330 &0.193 &0.342 &0.705  &0.899 &0.626 &0.783 &0.944&0.399&0.571&0.842  \\
		Multi-View&0.505 &0.543 &0.342 &0.202 &0.350 &0.729  &0.908 &0.662 &0.801 &0.951&0.421&0.601&0.861  \\
		DUA&0.551 &0.599 &0.421 &0.243 &0.421 &0.780  &- &0.752 &0.868 &0.962&0.501&0.700&0.921  \\
		DAM&0.550 &0.601 &0.427 &0.254 &0.410  &0.757 &0.938 &0.767 &0.874 &0.969&0.526&0.727&0.933 \\
		MRFN&0.571 &0.617 &0.448 &0.276 &0.435 &0.783  &0.945 &0.786 &0.886 &0.976&-&-&-  \\
		IOI& 0.573& 0.621&0.444 &0.269 &0.451 &0.786  &0.947 &0.796 &0.894 &0.974&0.563&0.768&0.950  \\
		\hline
		SMN  & 0.529 & 0.569 & 0.397 & 0.233 & 0.396 & 0.724  &0.926&0.726&0.847&0.961&0.453&0.654&0.886  \\
		MSN  & 0.587 & 0.632 & 0.470 & 0.295 & 0.452 & 0.788  &-&0.800&0.899&0.978&0.606&0.770&0.937   \\
		SA-BERT &  0.619&  0.659&  0.496&  0.313& 0.481 &   0.847&0.965&0.855&0.928&0.983&0.704&0.879&0.985   \\   
		\hline
		SMN+HCL  &0.575  & 0.620 &0.446  &0.281  &0.452  &0.807 &0.947&0.777&0.885&0.981&0.507&0.723&0.935   \\
		MSN+HCL &0.620  &0.668  &0.507  &0.321  &0.508  &0.841 &0.969&0.826&0.924&0.989&0.642&0.814&0.968    \\
		SA-BERT+HCL &\textbf{0.639}  &\textbf{0.681}  &\textbf{0.514}  &\textbf{0.330}  &\textbf{0.531}  &\textbf{0.858}   &\textbf{0.977}&\textbf{0.867}&\textbf{0.940}&\textbf{0.992}&\textbf{0.721}&\textbf{0.896}&\textbf{0.993}   \\
		\hlinewd{0.75pt}
	\end{tabular}}
	\caption{Experimental results of different models trained with our approach on Douban, Ubuntu, and E-Commerce datasets. All results acquired using HCL outperforms the original results with a significance level $p\textup{-value} < 0.01$.
	}
	\label{tb:main_result}
\end{table*}

\paragraph{Curriculum Learning.} Curriculum Learning \cite{DBLP:conf/icml/BengioLCW09} is reminiscent of the cognitive process of human being. Its core idea is first learning easier concepts and then gradually transitioning to more complex concepts based on some predefined learning schemes. Curriculum learning (CL) has demonstrated its benefits in various machine learning tasks \cite{DBLP:conf/naacl/SpitkovskyAJ10,DBLP:conf/cvpr/IlgMSKDB17,DBLP:conf/bmvc/LiZHXK17,DBLP:conf/aaai/SvetlikLSSWS17,DBLP:conf/ijcai/LiuH0018,DBLP:conf/naacl/PlataniosSNPM19}. Recently, \citet{DBLP:conf/ecir/PenhaH20} employed the idea of CL to tackle the response selection task. However, they only apply curriculum learning for the positive-side response selection, while ignoring the diversity of the negative responses.

\section{Experiment Setups}
\subsection{Datasets and Evaluation Metrics}
We test our approach on three benchmark datasets.

\paragraph{Douban Dataset.} This dataset \cite{DBLP:conf/acl/WuWXZL17} consists of multi-turn Chinese conversation data crawled from Douban group\footnote{https://www.douban.com/group}. The size of training, validation and test set are 500k, 25k and 1k. In the test set, each dialogue context is paired with 10 candidate responses. Following previous works, we report the results of Mean Average Precision (MAP), Mean Reciprocal Rank (MRR) and Precision at Position 1 ($\textup{P@1}$). In addition, we also report the results of $\textup{R}_{10}@\textup{1}$, $\textup{R}_{10}@\textup{2}$, $\textup{R}_{10}\textup{@5}$, where $\textup{R}_{n}\textup{@}k$ means recall at position $k$ in $n$ candidates. 

\paragraph{Ubuntu Dataset.} This dataset \cite{DBLP:conf/sigdial/LowePSP15} contains multi-turn dialogues collected from chat logs of the Ubuntu Forum. The training, validation and test size are 500k, 50k and 50k. Each dialogue context is paired with 10 response candidates. Following previous studies, we use $\textup{R}_{2}@\textup{1}$, $\textup{R}_{10}@\textup{1}$, $\textup{R}_{10}@\textup{2}$ and $\textup{R}_{10}@\textup{5}$ as evaluation metrics.

\paragraph{E-Commerce Dataset.} This dataset \cite{DBLP:conf/coling/ZhangLZZL18} consists of Chinese conversations between customers and customer service staff from Taobao\footnote{www.taobao.com}. The size of training, validation and test set are 500k, 25k and 1k. In the test set, each dialogue context is paired with 10 candidate responses. $\textup{R}_{n}@\textup{k}$ are employed as the evaluation metrics. 

\subsection{Baseline Models}
In the experiments, we compare our approach with the following models that can be summarized into three categories. 

\paragraph{Single-turn Matching Models.} This type of models treats all dialogue context as a single long utterance and then measures the relevance score between the context and response candidates, including RNN \cite{DBLP:conf/sigdial/LowePSP15}, CNN \cite{DBLP:conf/sigdial/LowePSP15}, LSTM \cite{DBLP:conf/sigdial/LowePSP15}, Bi-LSTM \cite{DBLP:journals/corr/KadlecSK15},  MV-LSTM \cite{DBLP:conf/ijcai/WanLXGPC16} and Match-LSTM \cite{DBLP:conf/naacl/WangJ16}.

\paragraph{Multi-turn Matching Models.} Instead of treating the dialogue context as one single utterance, these models aggregate information from different utterances in more sophisticated ways, including DL2R \cite{DBLP:conf/sigir/YanSW16}, Multi-View \cite{DBLP:conf/emnlp/ZhouDWZYTLY16}, DUA \cite{DBLP:conf/coling/ZhangLZZL18}, DAM \cite{DBLP:conf/acl/WuLCZDYZL18}, 
MRFN \cite{10.1145/3289600.3290985}, IOI \cite{DBLP:conf/acl/TaoWXHZY19}, SMN \cite{DBLP:conf/acl/WuWXZL17} and MSN \cite{DBLP:conf/emnlp/YuanZLLZHH19}. 

\paragraph{BERT-based Matching Models.} Given the recent advances of pre-trained language models \cite{DBLP:conf/naacl/DevlinCLT19}, \citet{DBLP:conf/cikm/GuLLLSWZ20} proposed the SA-BERT model which adapts BERT for the task of response selection and it is the current state-of-the-art model on the Douban and Ubuntu dataset. 

\begin{table*}[tb]
    \small
	\centering  
	\renewcommand{\arraystretch}{1.15}
	\scalebox{0.9}{
	\begin{tabular}{ccccccccccc}
		\hlinewd{0.75pt}
		\multirow{2}{*}{CC}&\multirow{2}{*}{IC}&\multicolumn{3}{c}{\textbf{SMN}}&\multicolumn{3}{c}{\textbf{MSN}}&\multicolumn{3}{c}{\textbf{SA-BERT}}\\
		\cmidrule(lr){3-5}
		\cmidrule(lr){6-8}
		\cmidrule(lr){9-11}
		&&$\rm P$@$1$ & $\rm R_{10}$@$1$&$\rm R_{10}$@$2$&$\rm P$@$1$ & $\rm R_{10}$@$1$&$\rm R_{10}$@$2$&$\rm P$@$1$ & $\rm R_{10}$@$1$&$\rm R_{10}$@$2$\\
		\hline
        $\times$&$\times$&0.402&0.238&0.410&0.474&0.298&0.462&0.499&0.315&0.493\\
        \hline
        \checkmark&$\times$&0.422&0.253&0.429&0.482&0.305&0.479&0.504&0.320&0.511\\
        \hline
        $\times$&\checkmark&0.441&0.271&0.444&0.499&0.315&0.492&0.511&0.325&0.524\\
        \hline
        \checkmark&\checkmark&\textbf{0.446}&\textbf{0.281}&\textbf{0.452}&\textbf{0.507}&\textbf{0.321}&\textbf{0.508}&\textbf{0.514}&\textbf{0.330}&\textbf{0.531}\\
		\hlinewd{0.75pt}
	\end{tabular}}
    \caption{Ablation study on Douban dataset using different combinations of the proposed curriculum strategies.}
	\label{tb:ablation_study}
\end{table*}

\begin{table*}[tb]
    \small
	\centering  
	\renewcommand{\arraystretch}{1.15}
	\scalebox{0.85}{
	\begin{tabular}{ccccccccccc}
	    \hlinewd{0.75pt}
	    \multicolumn{1}{c}{\multirow{2}{*}{Model}}&\multicolumn{1}{c}{\multirow{2}{*}{Strategy}}& \multicolumn{5}{c}{\textbf{Douban}}&\multicolumn{4}{c}{\textbf{Ubuntu}}\\
	    \cmidrule(lr){3-7}
	    \cmidrule(lr){8-11}
	    && $\rm {MAP}$ & $\rm MRR$&$\rm P$@$1$ & $\rm R_{10}$@$1$ & $\rm R_{10}$@$2$&$\rm R_{2}$@$1$&$\rm R_{10}$@$1$ & $\rm R_{10}$@$2$ & $\rm R_{10}$@$5$\\
	    \hline
	    \multirow{4}{*}{SMN}&Semi&0.554&0.605&0.425&0.253&0.412&0.934&0.762&0.865&0.967\\
	    &CIR$\ddag$&0.561&0.611&0.432&0.267&0.433&0.935&0.760&0.870&0.963\\
		&Gray&0.564&0.615&0.443&0.271&0.439&0.938&0.765&0.873&0.969\\
		&HCL&\textbf{0.575}&\textbf{0.620}&\textbf{0.446}&\textbf{0.281}&\textbf{0.452}&\textbf{0.947}&\textbf{0.777}&\textbf{0.885}&\textbf{0.981}\\
		\hline
	    \multirow{4}{*}{MSN}&Semi$\ddag$&0.591&0.638&0.473&0.301&0.461&0.952&0.804&0.903&0.983\\
	    &CIR$\ddag$&0.595&0.640&0.472&0.304&0.466&0.955&0.808&0.910&0.985\\
		&Gray&0.599&0.645&0.476&0.308&0.468&0.958&0.812&0.911&0.987\\
		&HCL&\textbf{0.620}&\textbf{0.668}&\textbf{0.507}&\textbf{0.321}&\textbf{0.508}&\textbf{0.969}&\textbf{0.826}&\textbf{0.924}&\textbf{0.989}\\
		\hline
	    \multirow{4}{*}{SA-BERT}&Semi${\ddag}$&0.623&0.664&0.500&0.317&0.490&0.968&0.858&0.931&0.989\\
	    &CIR$\ddag$&0.624&0.666&0.503&0.318&0.497&0.969&0.860&0.935&0.990\\
		&Gray$\ddag$&0.628&0.670&0.503&0.320&0.503&0.970&0.861&0.934&0.991\\
		&HCL&\textbf{0.639}&\textbf{0.681}&\textbf{0.514}&\textbf{0.330}&\textbf{0.531}&\textbf{0.977}&\textbf{0.867}&\textbf{0.940}&\textbf{0.992}\\
		\hlinewd{0.75pt}
	\end{tabular}}
    \caption{Comparisons on Douban and Ubuntu datasets using different training strategies on various models. 
    Results marked with $\ddag$ are from our runs with their released code. 
    }
	\label{tb:learning_comparison}
\end{table*}

\subsection{Implementation Details} 
For all experiments, we set the value of $p_{cc}(0)$ in the corpus-level pacing function $p_{cc}(t)$ as $0.3$, meaning that all models start training with the context-response pairs whose corpus-level difficulty is lower than $0.3$. For the instance-level pacing function $p_{ic}(t)$, the value of $k_T$ is set as $3$, meaning that, after IC is completed, the negative responses of each training instance are sampled from the top-$10^3$ relevant responses. In the experiments, each matching model is trained for $40,000$ steps with a batch size of 128, and we set the $T$ in both $p_{cc}(t)$ and $p_{ic}(t)$ as half of the total training steps, i.e., $T=20,000$. To build the context and response encoders in the ranking model $G(\cdot, \cdot)$, we use a $3$-layer transformers with a hidden size of $256$. We select two representative models (SMN and MSN) along with the state-of-the-art SA-BERT to test the proposed learning framework. To better simulate the true testing environment, the number of negative responses ($m$ in Eq. \eqref{eq:matching_model}) is set to be 5.

\section{Result and Analysis}
\subsection{Main Results}
Table \ref{tb:main_result} shows the results on Douban, Ubuntu, and E-Commerce datasets, where X+HCL means training the model X with the proposed learning HCL. We can see that HCL significantly improves the performance of all three matching models in terms of all evaluation metrics, showing the robustness and universality of our approach. We also observe that, by training with HCL, a model (MSN) without using pre-trained language model can even surpass the state-of-the-art model using pre-trained language model (SA-BERT) on Douban dataset. These results suggest that, while the training strategy is under-explored in previous studies, it could be very decisive for building a competent response selection model.

\subsection{Effect of CC and IC} 
To reveal the individual effects of CC and IC, we train different models on Douban dataset by removing either CC or IC. The experimental results are shown in Table \ref{tb:ablation_study}, from which we see that both CC and IC make positive contributions to the overall performance when used alone. Only utilizing IC leads to larger improvements than only using CC. This observation suggests that the ability of identifying the mismatching information is a more important factor for the model to achieve its optimal performance. However, the optimal performance is achieved when CC and IC are combined, indicating that CC and IC are complementary to each other.

\subsection{Contrast to Existing Learning Strategies}
Next, we compare our approach with other learning strategies proposed recently \cite{li-etal-2019-sampling,DBLP:conf/ecir/PenhaH20,lin2020world}. We use Semi, CIR, and Gray to denote the approaches in \citet{li-etal-2019-sampling}, \citet{DBLP:conf/ecir/PenhaH20}, and \citet{lin2020world} respectively, where Gray is the current state of the art. We conduct experiments on Douban and Ubuntu datasets and the experimental results of three matching models are listed in Table \ref{tb:learning_comparison}. 
From the results, we can see that our approach consistently outperforms other learning strategies in all settings. The performance gains of our approach are even more remarkable given its simplicity; it does not require running additional generation models \cite{lin2020world} or re-scoring negative samples at different epochs \cite{li-etal-2019-sampling}.

\begin{figure}[!t] 
	\centering    
	\setlength{\abovecaptionskip}{3pt}
\includegraphics[width=0.42\textwidth]{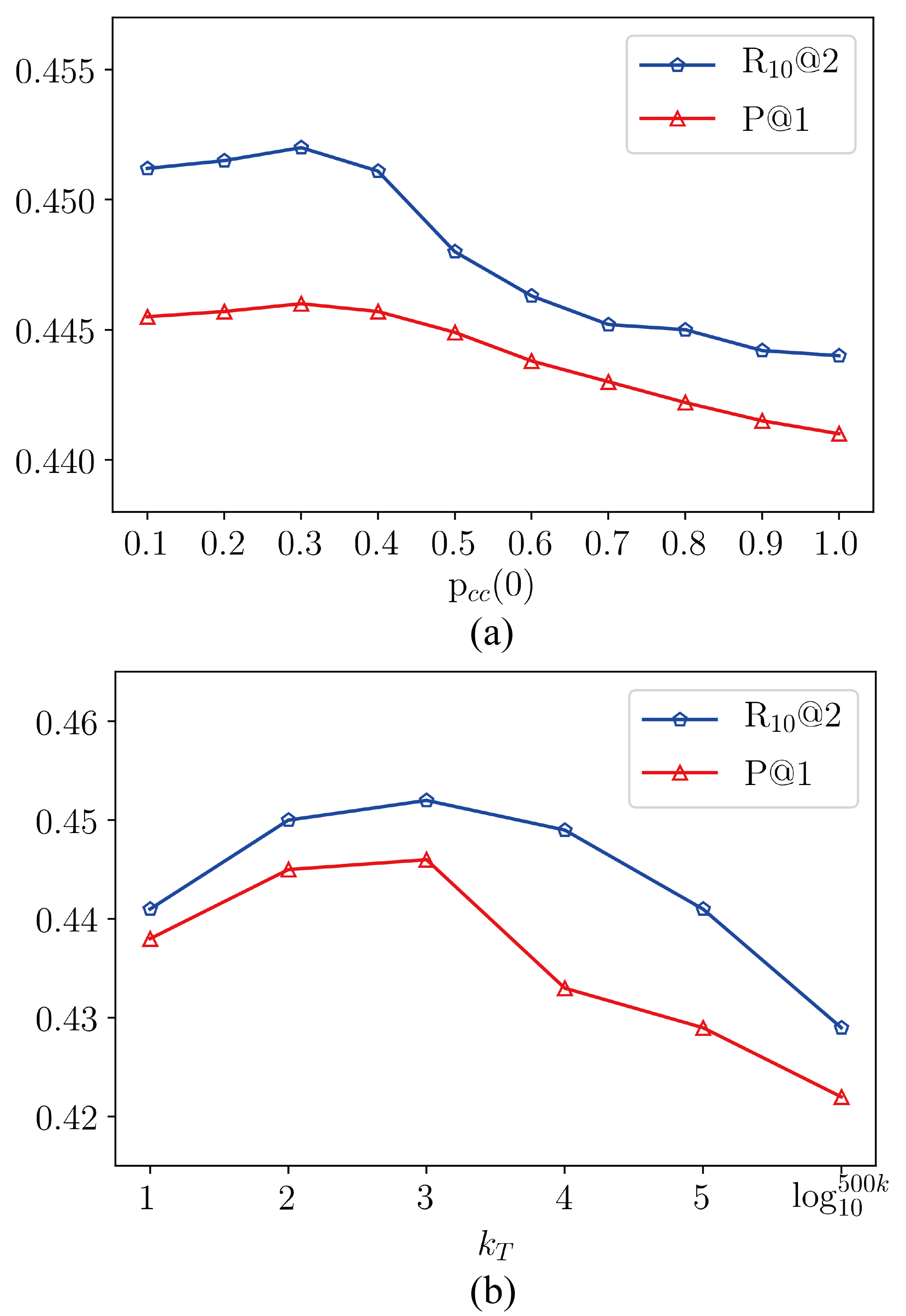}
	\caption{Plots illustrating the effect of curriculum hyper-parameters, (a) $p_{cc}(0)$ and (b) $k_T$, on the SMN model performance in Douban dataset.}
    \label{fig:hyperparameter_curve}
\end{figure}
\subsection{Further Analysis on HCL}
In this part, we study how the key hyper-parameters affect the performance of HCL, including the initial difficulty of CC, $p_{cc}(0)$, and the curriculum length of IC, $k_T$.\footnote{Our experiments show that other hyper-parameter settings have little impact on the model performance.} In addition, we also investigate the effect of different ranking model choices.
\paragraph{Initial Difficulty of CC.} 
We run sensitivity analysis experiments on Douban dataset with the SMN model by tuning $p_{cc}(0)$ in the corpus-level pacing function $p_{cc}(t)$. The results of $\rm P$@$1$ and $\rm R_{10}$@$2$ in terms of $p_{cc}(0)$ and $k_T$ are shown in Figure \ref{fig:hyperparameter_curve}(a). We observe that when $p_{cc}(0)$ is small (i.e., $p_{cc}(0)\leq 0.3$), the model performances are relatively similar. When $p_{cc}(0)$ approaches to 1.0, the results drop significantly. It concurs with our expectation that, in CC, the model should start learning with training context-response pairs of lower difficulty. Once $p_{cc}(0)$ becomes 1.0, the CC is disabled, resulting the lowest model performances.

\paragraph{Curriculum Length of IC.} Similair to $p_{cc}(0)$, we also run sensitivity analysis experiments by tuning $k_T$ in the instance-level pacing function $p_{ic}(t)$ and Figure \ref{fig:hyperparameter_curve}(b) shows the results. We observe that a too small or too large $K_T$ results in performance degradation. When $k_T$ is too small, after IC is completed, the negative examples are only sampled from a very small subset of the training data that consists of responses with high relevance. In this case, the sampled responses might be false negatives that should be deemed as positive cases. Thus, learning to treat those responses as true negatives could harm the model performance. On the other hand, as $k_T$ increases, the effect of IC becomes less obvious. When $k_T=\log_{10}^{500k}$ ($|\mathcal{D}|=500k$), IC is completely disabled, leading to the further decrease of model performances. 

\begin{table}
    \small
	\centering  
	\renewcommand{\arraystretch}{1.15}
	\scalebox{0.86}{
	\begin{tabular}{c|c|ccc}
	    \hlinewd{0.75pt}
	    \textbf{Ranking Model}&Model&$\rm P$@$1$ & $\rm R_{10}$@$1$ & $\rm R_{10}$@$2$\\
	    \hline
	    \multirow{4}{*}{Transformers}&Ranking Model&0.400&0.253&0.416\\\cline{2-5}
	    &SMN&0.446&\textbf{0.281}&0.452\\
		&MSN&\textbf{0.507}&0.321&\textbf{0.508}\\
		&SA-BERT&\textbf{0.514}&\textbf{0.330}&0.531\\
		\hline
	    \multirow{4}{*}{BiLSTM}&Ranking Model&0.377&0.227&0.393\\\cline{2-5}
	    &SMN&0.438&0.273&0.441\\
		&MSN&0.491&0.313&0.487\\
		&SA-BERT&0.507&0.323&0.513\\
		\hline
	    \multirow{4}{*}{BERT-base}&Ranking Model&0.437&0.275&0.443\\\cline{2-5}
	    &SMN&\textbf{0.451}&0.279&\textbf{0.457}\\
		&MSN&\textbf{0.507}&\textbf{0.323}&0.507\\
		&SA-BERT&0.511&0.329&\textbf{0.53}5\\
		\hlinewd{0.75pt}
	\end{tabular}}
    \caption{Comparisons of different ranking model architectures. Best results of each matching model are \textbf{bold-faced}. The ``Ranking Model'' rows represent the performances of different ranking models.}
	\label{tb:ranker_choice}
\end{table}

\paragraph{Ranking Model Architecture.}
\label{sec:ranker_choice}
Lastly, we examine the effect of the choice of the ranking model architecture. We build two ranking model variants by replacing the Transformers module $E_c(\cdot)$ and $E_r(\cdot)$ in Eq. \eqref{eq:similarity} with other modules. For the first case, we use $3$-layer BiLSTM with a hidden size of 256. For the second one, we use BERT-base \cite{DBLP:conf/naacl/DevlinCLT19} model. Then, we train the matching models using the proposed HCL but with different ranking models as the scoring basis.

The results on Douban dataset are shown in Table \ref{tb:ranker_choice}. We first compare the performance of different ranking models by directly using them to select the best response. The results are shown in the ``Ranking Model" row of Table \ref{tb:ranker_choice}. Among all three variants, BERT performs the best but it is still less accurate than these sophisticated matching models.  
Second, we study the effect of different ranking models on the matching model performance. We see that, for different matching models, Transformers and BERT perform comparably but the results from BiLSTM are much worse. This further leads to a conclusion that, while the choice of ranking model does have impact on the overall results, the improvement of the ranking model  does not necessarily lead to the improvement of matching models once the ranking model achieves certain accuracy.

\section{Conclusion}
In this work, we propose a novel hierarchical curriculum learning framework for training response selection models for multi-turn conversations. During training, the proposed framework simultaneously employs corpus-level and instance-level curricula to dynamically select suitable training data based on the state of the learning process. Extensive experiments and analysis on two benchmark datasets show that our approach can significantly improve the performance of various strong matching models on all evaluation metrics. Our code, models and other related resources can be found in  \url{https://github.com/yxuansu/HCL}

\section*{Acknowledgments}
The authors wish to thank Jialu Xu and Sihui Wang for their insightful discussions and support. Many thanks to our anonymous reviewers for their suggestions and comments.

\section*{Ethical Statement}
We honor and support the ACL code of Ethics. Dialogue response selection aims to build a retrieval-based dialogue system which better interacts with users. The selection of the best response does not involve any bias towards to the participants. All datasets used in this work are from previously published works, and in our view, do not have any attached privacy or ethical issues.

\bibliographystyle{acl_natbib}
\bibliography{anthology,acl2021}

\begin{thebibliography}{39}
\expandafter\ifx\csname natexlab\endcsname\relax\def\natexlab#1{#1}\fi

\bibitem[{Bengio et~al.(2009)Bengio, Louradour, Collobert, and
  Weston}]{DBLP:conf/icml/BengioLCW09}
Yoshua Bengio, J{\'{e}}r{\^{o}}me Louradour, Ronan Collobert, and Jason Weston.
  2009.
\newblock \href {https://doi.org/10.1145/1553374.1553380} {Curriculum
  learning}.
\newblock In \emph{Proceedings of the 26th Annual International Conference on
  Machine Learning, {ICML} 2009, Montreal, Quebec, Canada, June 14-18, 2009},
  pages 41--48.

\bibitem[{Devlin et~al.(2019)Devlin, Chang, Lee, and
  Toutanova}]{DBLP:conf/naacl/DevlinCLT19}
Jacob Devlin, Ming{-}Wei Chang, Kenton Lee, and Kristina Toutanova. 2019.
\newblock \href {https://doi.org/10.18653/v1/n19-1423} {{BERT:} pre-training of
  deep bidirectional transformers for language understanding}.
\newblock In \emph{Proceedings of the 2019 Conference of the North American
  Chapter of the Association for Computational Linguistics: Human Language
  Technologies, {NAACL-HLT} 2019, Minneapolis, MN, USA, June 2-7, 2019, Volume
  1 (Long and Short Papers)}, pages 4171--4186. Association for Computational
  Linguistics.

\bibitem[{Elman(1990)}]{DBLP:journals/cogsci/Elman90}
Jeffrey~L. Elman. 1990.
\newblock \href {https://doi.org/10.1207/s15516709cog1402\_1} {Finding
  structure in time}.
\newblock \emph{Cogn. Sci.}, 14(2):179--211.

\bibitem[{Feng et~al.(2019)Feng, Tao, Wu, Feng, Zhao, and
  Yan}]{DBLP:conf/acl/FengTWFZY19}
Jiazhan Feng, Chongyang Tao, Wei Wu, Yansong Feng, Dongyan Zhao, and Rui Yan.
  2019.
\newblock \href {https://doi.org/10.18653/v1/p19-1370} {Learning a matching
  model with co-teaching for multi-turn response selection in retrieval-based
  dialogue systems}.
\newblock In \emph{Proceedings of the 57th Conference of the Association for
  Computational Linguistics, {ACL} 2019, Florence, Italy, July 28- August 2,
  2019, Volume 1: Long Papers}, pages 3805--3815. Association for Computational
  Linguistics.

\bibitem[{Gu et~al.(2020)Gu, Li, Liu, Ling, Su, Wei, and
  Zhu}]{DBLP:conf/cikm/GuLLLSWZ20}
Jia{-}Chen Gu, Tianda Li, Quan Liu, Zhen{-}Hua Ling, Zhiming Su, Si~Wei, and
  Xiaodan Zhu. 2020.
\newblock \href {https://doi.org/10.1145/3340531.3412330} {Speaker-aware {BERT}
  for multi-turn response selection in retrieval-based chatbots}.
\newblock In \emph{{CIKM} '20: The 29th {ACM} International Conference on
  Information and Knowledge Management, Virtual Event, Ireland, October 19-23,
  2020}, pages 2041--2044. {ACM}.

\bibitem[{Gu et~al.(2019)Gu, Ling, and Liu}]{DBLP:conf/cikm/GuLL19}
Jia{-}Chen Gu, Zhen{-}Hua Ling, and Quan Liu. 2019.
\newblock \href {https://doi.org/10.1145/3357384.3358140} {Interactive matching
  network for multi-turn response selection in retrieval-based chatbots}.
\newblock In \emph{Proceedings of the 28th {ACM} International Conference on
  Information and Knowledge Management, {CIKM} 2019, Beijing, China, November
  3-7, 2019}, pages 2321--2324.

\bibitem[{Hochreiter and Schmidhuber(1997)}]{DBLP:journals/neco/HochreiterS97}
Sepp Hochreiter and J{\"{u}}rgen Schmidhuber. 1997.
\newblock \href {https://doi.org/10.1162/neco.1997.9.8.1735} {Long short-term
  memory}.
\newblock \emph{Neural Comput.}, 9(8):1735--1780.

\bibitem[{Ilg et~al.(2017)Ilg, Mayer, Saikia, Keuper, Dosovitskiy, and
  Brox}]{DBLP:conf/cvpr/IlgMSKDB17}
Eddy Ilg, Nikolaus Mayer, Tonmoy Saikia, Margret Keuper, Alexey Dosovitskiy,
  and Thomas Brox. 2017.
\newblock \href {https://doi.org/10.1109/CVPR.2017.179} {Flownet 2.0: Evolution
  of optical flow estimation with deep networks}.
\newblock In \emph{2017 {IEEE} Conference on Computer Vision and Pattern
  Recognition, {CVPR} 2017, Honolulu, HI, USA, July 21-26, 2017}, pages
  1647--1655. {IEEE} Computer Society.

\bibitem[{Kadlec et~al.(2015)Kadlec, Schmid, and
  Kleindienst}]{DBLP:journals/corr/KadlecSK15}
Rudolf Kadlec, Martin Schmid, and Jan Kleindienst. 2015.
\newblock \href {http://arxiv.org/abs/1510.03753} {Improved deep learning
  baselines for ubuntu corpus dialogs}.
\newblock \emph{CoRR}, abs/1510.03753.

\bibitem[{Karpukhin et~al.(2020)Karpukhin, Oguz, Min, Lewis, Wu, Edunov, Chen,
  and Yih}]{DBLP:conf/emnlp/KarpukhinOMLWEC20}
Vladimir Karpukhin, Barlas Oguz, Sewon Min, Patrick S.~H. Lewis, Ledell Wu,
  Sergey Edunov, Danqi Chen, and Wen{-}tau Yih. 2020.
\newblock \href {https://www.aclweb.org/anthology/2020.emnlp-main.550/} {Dense
  passage retrieval for open-domain question answering}.
\newblock In \emph{Proceedings of the 2020 Conference on Empirical Methods in
  Natural Language Processing, {EMNLP} 2020, Online, November 16-20, 2020},
  pages 6769--6781. Association for Computational Linguistics.

\bibitem[{Kingma and Ba(2015)}]{DBLP:journals/corr/KingmaB14}
Diederik~P. Kingma and Jimmy Ba. 2015.
\newblock \href {http://arxiv.org/abs/1412.6980} {Adam: {A} method for
  stochastic optimization}.
\newblock In \emph{3rd International Conference on Learning Representations,
  {ICLR} 2015, San Diego, CA, USA, May 7-9, 2015, Conference Track
  Proceedings}.

\bibitem[{Kollar et~al.(2018)Kollar, Berry, Stuart, Owczarzak, Chung, Mathias,
  Kayser, Snow, and Matsoukas}]{DBLP:conf/naacl/KollarBSOCMKSM18}
Thomas Kollar, Danielle Berry, Lauren Stuart, Karolina Owczarzak, Tagyoung
  Chung, Lambert Mathias, Michael Kayser, Bradford Snow, and Spyros Matsoukas.
  2018.
\newblock \href {https://doi.org/10.18653/v1/n18-3022} {The alexa meaning
  representation language}.
\newblock In \emph{Proceedings of the 2018 Conference of the North American
  Chapter of the Association for Computational Linguistics: Human Language
  Technologies, {NAACL-HLT} 2018, New Orleans, Louisiana, USA, June 1-6, 2018,
  Volume 3 (Industry Papers)}, pages 177--184.

\bibitem[{Li et~al.(2019)Li, Tao, Wu, Feng, Zhao, and
  Yan}]{li-etal-2019-sampling}
Jia Li, Chongyang Tao, Wei Wu, Yansong Feng, Dongyan Zhao, and Rui Yan. 2019.
\newblock \href {https://doi.org/10.18653/v1/D19-1128} {Sampling matters! an
  empirical study of negative sampling strategies for learning of matching
  models in retrieval-based dialogue systems}.
\newblock In \emph{Proceedings of the 2019 Conference on Empirical Methods in
  Natural Language Processing and the 9th International Joint Conference on
  Natural Language Processing (EMNLP-IJCNLP)}, pages 1291--1296, Hong Kong,
  China. Association for Computational Linguistics.

\bibitem[{Li et~al.(2017)Li, Zhu, Huang, Xu, and Kuo}]{DBLP:conf/bmvc/LiZHXK17}
Siyang Li, Xiangxin Zhu, Qin Huang, Hao Xu, and C.{-}C.~Jay Kuo. 2017.
\newblock \href {https://www.dropbox.com/s/pfhxh9a9qnk7nx2/0052.pdf?dl=1}
  {Multiple instance curriculum learning for weakly supervised object
  detection}.
\newblock In \emph{British Machine Vision Conference 2017, {BMVC} 2017, London,
  UK, September 4-7, 2017}. {BMVA} Press.

\bibitem[{Lin et~al.(2020)Lin, Cai, Wang, Liu, Zheng, and Shi}]{lin2020world}
Zibo Lin, Deng Cai, Yan Wang, Xiaojiang Liu, Hai-Tao Zheng, and Shuming Shi.
  2020.
\newblock \href {http://arxiv.org/abs/2004.02421} {The world is not binary:
  Learning to rank with grayscale data for dialogue response selection}.

\bibitem[{Liu et~al.(2018)Liu, He, Liu, and Zhao}]{DBLP:conf/ijcai/LiuH0018}
Cao Liu, Shizhu He, Kang Liu, and Jun Zhao. 2018.
\newblock \href {https://doi.org/10.24963/ijcai.2018/587} {Curriculum learning
  for natural answer generation}.
\newblock In \emph{Proceedings of the Twenty-Seventh International Joint
  Conference on Artificial Intelligence, {IJCAI} 2018, July 13-19, 2018,
  Stockholm, Sweden}, pages 4223--4229. ijcai.org.

\bibitem[{Lowe et~al.(2015)Lowe, Pow, Serban, and
  Pineau}]{DBLP:conf/sigdial/LowePSP15}
Ryan Lowe, Nissan Pow, Iulian Serban, and Joelle Pineau. 2015.
\newblock \href {https://doi.org/10.18653/v1/w15-4640} {The ubuntu dialogue
  corpus: {A} large dataset for research in unstructured multi-turn dialogue
  systems}.
\newblock In \emph{Proceedings of the {SIGDIAL} 2015 Conference, The 16th
  Annual Meeting of the Special Interest Group on Discourse and Dialogue, 2-4
  September 2015, Prague, Czech Republic}, pages 285--294. The Association for
  Computer Linguistics.

\bibitem[{Lu et~al.(2019)Lu, Zhang, Xie, Ling, Zhou, and
  Xu}]{DBLP:conf/acl/LuZXLZX19}
Junyu Lu, Chenbin Zhang, Zeying Xie, Guang Ling, Tom~Chao Zhou, and Zenglin Xu.
  2019.
\newblock \href {https://doi.org/10.18653/v1/p19-1006} {Constructing
  interpretive spatio-temporal features for multi-turn responses selection}.
\newblock In \emph{Proceedings of the 57th Conference of the Association for
  Computational Linguistics, {ACL} 2019, Florence, Italy, July 28- August 2,
  2019, Volume 1: Long Papers}, pages 44--50.

\bibitem[{Penha and Hauff(2020)}]{DBLP:conf/ecir/PenhaH20}
Gustavo Penha and Claudia Hauff. 2020.
\newblock \href {https://doi.org/10.1007/978-3-030-45439-5\_46} {Curriculum
  learning strategies for {IR}}.
\newblock In \emph{Advances in Information Retrieval - 42nd European Conference
  on {IR} Research, {ECIR} 2020, Lisbon, Portugal, April 14-17, 2020,
  Proceedings, Part {I}}, volume 12035 of \emph{Lecture Notes in Computer
  Science}, pages 699--713. Springer.

\bibitem[{Platanios et~al.(2019)Platanios, Stretcu, Neubig, P{\'{o}}czos, and
  Mitchell}]{DBLP:conf/naacl/PlataniosSNPM19}
Emmanouil~Antonios Platanios, Otilia Stretcu, Graham Neubig, Barnab{\'{a}}s
  P{\'{o}}czos, and Tom~M. Mitchell. 2019.
\newblock \href {https://doi.org/10.18653/v1/n19-1119} {Competence-based
  curriculum learning for neural machine translation}.
\newblock In \emph{Proceedings of the 2019 Conference of the North American
  Chapter of the Association for Computational Linguistics: Human Language
  Technologies, {NAACL-HLT} 2019, Minneapolis, MN, USA, June 2-7, 2019, Volume
  1 (Long and Short Papers)}, pages 1162--1172. Association for Computational
  Linguistics.

\bibitem[{Ritter et~al.(2011)Ritter, Cherry, and
  Dolan}]{DBLP:conf/emnlp/RitterCD11}
Alan Ritter, Colin Cherry, and William~B. Dolan. 2011.
\newblock \href {https://www.aclweb.org/anthology/D11-1054/} {Data-driven
  response generation in social media}.
\newblock In \emph{Proceedings of the 2011 Conference on Empirical Methods in
  Natural Language Processing, {EMNLP} 2011, 27-31 July 2011, John McIntyre
  Conference Centre, Edinburgh, UK, {A} meeting of SIGDAT, a Special Interest
  Group of the {ACL}}, pages 583--593.

\bibitem[{Shum et~al.(2018)Shum, He, and
  Li}]{DBLP:journals/corr/abs-1801-01957}
Heung{-}Yeung Shum, Xiaodong He, and Di~Li. 2018.
\newblock \href {http://arxiv.org/abs/1801.01957} {From eliza to xiaoice:
  Challenges and opportunities with social chatbots}.
\newblock \emph{CoRR}, abs/1801.01957.

\bibitem[{Spitkovsky et~al.(2010)Spitkovsky, Alshawi, and
  Jurafsky}]{DBLP:conf/naacl/SpitkovskyAJ10}
Valentin~I. Spitkovsky, Hiyan Alshawi, and Daniel Jurafsky. 2010.
\newblock \href {https://www.aclweb.org/anthology/N10-1116/} {From baby steps
  to leapfrog: How "less is more" in unsupervised dependency parsing}.
\newblock In \emph{Human Language Technologies: Conference of the North
  American Chapter of the Association of Computational Linguistics,
  Proceedings, June 2-4, 2010, Los Angeles, California, {USA}}, pages 751--759.
  The Association for Computational Linguistics.

\bibitem[{Su et~al.(2020)Su, Wang, Baker, Cai, Liu, Korhonen, and
  Collier}]{DBLP:journals/corr/abs-2004-02214}
Yixuan Su, Yan Wang, Simon Baker, Deng Cai, Xiaojiang Liu, Anna Korhonen, and
  Nigel Collier. 2020.
\newblock \href {http://arxiv.org/abs/2004.02214} {Prototype-to-style: Dialogue
  generation with style-aware editing on retrieval memory}.
\newblock \emph{CoRR}, abs/2004.02214.

\bibitem[{Svetlik et~al.(2017)Svetlik, Leonetti, Sinapov, Shah, Walker, and
  Stone}]{DBLP:conf/aaai/SvetlikLSSWS17}
Maxwell Svetlik, Matteo Leonetti, Jivko Sinapov, Rishi Shah, Nick Walker, and
  Peter Stone. 2017.
\newblock \href {http://aaai.org/ocs/index.php/AAAI/AAAI17/paper/view/14961}
  {Automatic curriculum graph generation for reinforcement learning agents}.
\newblock In \emph{Proceedings of the Thirty-First {AAAI} Conference on
  Artificial Intelligence, February 4-9, 2017, San Francisco, California,
  {USA}}, pages 2590--2596. {AAAI} Press.

\bibitem[{Tan et~al.(2016)Tan, dos Santos, Xiang, and Zhou}]{tan2016lstmbased}
Ming Tan, Cicero dos Santos, Bing Xiang, and Bowen Zhou. 2016.
\newblock \href {http://arxiv.org/abs/1511.04108} {Lstm-based deep learning
  models for non-factoid answer selection}.

\bibitem[{Tao et~al.(2019{\natexlab{a}})Tao, Wu, Xu, Hu, Zhao, and
  Yan}]{10.1145/3289600.3290985}
Chongyang Tao, Wei Wu, Can Xu, Wenpeng Hu, Dongyan Zhao, and Rui Yan.
  2019{\natexlab{a}}.
\newblock Multi-representation fusion network for multi-turn response selection
  in retrieval-based chatbots.
\newblock In \emph{Proceedings of the Twelfth ACM International Conference on
  Web Search and Data Mining}, page 267–275, New York, NY, USA. Association
  for Computing Machinery.

\bibitem[{Tao et~al.(2019{\natexlab{b}})Tao, Wu, Xu, Hu, Zhao, and
  Yan}]{DBLP:conf/acl/TaoWXHZY19}
Chongyang Tao, Wei Wu, Can Xu, Wenpeng Hu, Dongyan Zhao, and Rui Yan.
  2019{\natexlab{b}}.
\newblock \href {https://doi.org/10.18653/v1/p19-1001} {One time of interaction
  may not be enough: Go deep with an interaction-over-interaction network for
  response selection in dialogues}.
\newblock In \emph{Proceedings of the 57th Conference of the Association for
  Computational Linguistics, {ACL} 2019, Florence, Italy, July 28- August 2,
  2019, Volume 1: Long Papers}, pages 1--11.

\bibitem[{Vaswani et~al.(2017)Vaswani, Shazeer, Parmar, Uszkoreit, Jones,
  Gomez, Kaiser, and Polosukhin}]{DBLP:conf/nips/VaswaniSPUJGKP17}
Ashish Vaswani, Noam Shazeer, Niki Parmar, Jakob Uszkoreit, Llion Jones,
  Aidan~N. Gomez, Lukasz Kaiser, and Illia Polosukhin. 2017.
\newblock \href {http://papers.nips.cc/paper/7181-attention-is-all-you-need}
  {Attention is all you need}.
\newblock In \emph{Advances in Neural Information Processing Systems 30: Annual
  Conference on Neural Information Processing Systems 2017, 4-9 December 2017,
  Long Beach, CA, {USA}}, pages 5998--6008.

\bibitem[{Wan et~al.(2016)Wan, Lan, Xu, Guo, Pang, and
  Cheng}]{DBLP:conf/ijcai/WanLXGPC16}
Shengxian Wan, Yanyan Lan, Jun Xu, Jiafeng Guo, Liang Pang, and Xueqi Cheng.
  2016.
\newblock \href {http://www.ijcai.org/Abstract/16/415} {Match-srnn: Modeling
  the recursive matching structure with spatial {RNN}}.
\newblock In \emph{Proceedings of the Twenty-Fifth International Joint
  Conference on Artificial Intelligence, {IJCAI} 2016, New York, NY, USA, 9-15
  July 2016}, pages 2922--2928. {IJCAI/AAAI} Press.

\bibitem[{Wang et~al.(2013)Wang, Lu, Li, and Chen}]{DBLP:conf/emnlp/WangLLC13}
Hao Wang, Zhengdong Lu, Hang Li, and Enhong Chen. 2013.
\newblock \href {https://www.aclweb.org/anthology/D13-1096/} {A dataset for
  research on short-text conversations}.
\newblock In \emph{Proceedings of the 2013 Conference on Empirical Methods in
  Natural Language Processing, {EMNLP} 2013, 18-21 October 2013, Grand Hyatt
  Seattle, Seattle, Washington, USA, {A} meeting of SIGDAT, a Special Interest
  Group of the {ACL}}, pages 935--945. {ACL}.

\bibitem[{Wang and Jiang(2016)}]{DBLP:conf/naacl/WangJ16}
Shuohang Wang and Jing Jiang. 2016.
\newblock \href {https://doi.org/10.18653/v1/n16-1170} {Learning natural
  language inference with {LSTM}}.
\newblock In \emph{{NAACL} {HLT} 2016, The 2016 Conference of the North
  American Chapter of the Association for Computational Linguistics: Human
  Language Technologies, San Diego California, USA, June 12-17, 2016}, pages
  1442--1451. The Association for Computational Linguistics.

\bibitem[{Wu et~al.(2018)Wu, Wu, Li, and Zhou}]{DBLP:conf/acl/WuwLZ18}
Yu~Wu, Wei Wu, Zhoujun Li, and Ming Zhou. 2018.
\newblock Learning matching models with weak supervision for response selection
  in retrieval-based chatbots.
\newblock In \emph{Proceedings of the 56th Annual Meeting of the Association
  for Computational Linguistics, {ACL} 2018, Melbourne, Australia, July 15-20,
  2018, Volume 2: Short Papers}, pages 420--425.

\bibitem[{Wu et~al.(2017)Wu, Wu, Xing, Zhou, and Li}]{DBLP:conf/acl/WuWXZL17}
Yu~Wu, Wei Wu, Chen Xing, Ming Zhou, and Zhoujun Li. 2017.
\newblock \href {https://doi.org/10.18653/v1/P17-1046} {Sequential matching
  network: {A} new architecture for multi-turn response selection in
  retrieval-based chatbots}.
\newblock In \emph{Proceedings of the 55th Annual Meeting of the Association
  for Computational Linguistics, {ACL} 2017, Vancouver, Canada, July 30 -
  August 4, Volume 1: Long Papers}, pages 496--505.

\bibitem[{Yan et~al.(2016)Yan, Song, and Wu}]{DBLP:conf/sigir/YanSW16}
Rui Yan, Yiping Song, and Hua Wu. 2016.
\newblock \href {https://doi.org/10.1145/2911451.2911542} {Learning to respond
  with deep neural networks for retrieval-based human-computer conversation
  system}.
\newblock In \emph{Proceedings of the 39th International {ACM} {SIGIR}
  conference on Research and Development in Information Retrieval, {SIGIR}
  2016, Pisa, Italy, July 17-21, 2016}, pages 55--64. {ACM}.

\bibitem[{Yuan et~al.(2019)Yuan, Zhou, Li, Lv, Zhu, Han, and
  Hu}]{DBLP:conf/emnlp/YuanZLLZHH19}
Chunyuan Yuan, Wei Zhou, Mingming Li, Shangwen Lv, Fuqing Zhu, Jizhong Han, and
  Songlin Hu. 2019.
\newblock \href {https://doi.org/10.18653/v1/D19-1011} {Multi-hop selector
  network for multi-turn response selection in retrieval-based chatbots}.
\newblock In \emph{Proceedings of the 2019 Conference on Empirical Methods in
  Natural Language Processing and the 9th International Joint Conference on
  Natural Language Processing, {EMNLP-IJCNLP} 2019, Hong Kong, China, November
  3-7, 2019}, pages 111--120. Association for Computational Linguistics.

\bibitem[{Zhang et~al.(2018)Zhang, Li, Zhu, Zhao, and
  Liu}]{DBLP:conf/coling/ZhangLZZL18}
Zhuosheng Zhang, Jiangtong Li, Pengfei Zhu, Hai Zhao, and Gongshen Liu. 2018.
\newblock \href {https://www.aclweb.org/anthology/C18-1317/} {Modeling
  multi-turn conversation with deep utterance aggregation}.
\newblock In \emph{Proceedings of the 27th International Conference on
  Computational Linguistics, {COLING} 2018, Santa Fe, New Mexico, USA, August
  20-26, 2018}, pages 3740--3752. Association for Computational Linguistics.

\bibitem[{Zhou et~al.(2016)Zhou, Dong, Wu, Zhao, Yu, Tian, Liu, and
  Yan}]{DBLP:conf/emnlp/ZhouDWZYTLY16}
Xiangyang Zhou, Daxiang Dong, Hua Wu, Shiqi Zhao, Dianhai Yu, Hao Tian, Xuan
  Liu, and Rui Yan. 2016.
\newblock \href {https://doi.org/10.18653/v1/d16-1036} {Multi-view response
  selection for human-computer conversation}.
\newblock In \emph{Proceedings of the 2016 Conference on Empirical Methods in
  Natural Language Processing, {EMNLP} 2016, Austin, Texas, USA, November 1-4,
  2016}, pages 372--381. The Association for Computational Linguistics.

\bibitem[{Zhou et~al.(2018)Zhou, Li, Dong, Liu, Chen, Zhao, Yu, and
  Wu}]{DBLP:conf/acl/WuLCZDYZL18}
Xiangyang Zhou, Lu~Li, Daxiang Dong, Yi~Liu, Ying Chen, Wayne~Xin Zhao, Dianhai
  Yu, and Hua Wu. 2018.
\newblock \href {https://doi.org/10.18653/v1/P18-1103} {Multi-turn response
  selection for chatbots with deep attention matching network}.
\newblock In \emph{Proceedings of the 56th Annual Meeting of the Association
  for Computational Linguistics, {ACL} 2018, Melbourne, Australia, July 15-20,
  2018, Volume 1: Long Papers}, pages 1118--1127.

\end{thebibliography}

\clearpage
\appendix
\section{Ranking Model Training}
Here we provide more details on how to train the neural ranking model $G(\cdot, \cdot)$ that serves as the scoring basis in the proposed HCL framework.

\paragraph{Modelling.} Given a dialogue context $c$ and a response $r$, their context-response relevance score is defined as 
\begin{align}
\begin{split}
    G(c, r) &= E_c(c)^T E_r(r).
\end{split}
\end{align}
Note that, the context $c$ is a long sequence which is acquired by concatenating all utterances in the dialogue context. The $E_c(c)$ and $E_r(r)$ are the context and response encoder. The context encoder $E_c(\cdot)$ takes the token sequence $c=c_0,...,c_{|c|}$ and returns the context representation $E_c(c)$ by taking the output state corresponds to the last token $c_{|c|}$. The same operation is applied when computing the response representation $E_r(r)$. In practice, the choice of $E(\cdot)$ could be any sequence model, e.g., LSTM \cite{DBLP:journals/neco/HochreiterS97}, RNN \cite{DBLP:journals/cogsci/Elman90},  Transformers \cite{DBLP:conf/nips/VaswaniSPUJGKP17}, and BERT \cite{DBLP:conf/naacl/DevlinCLT19}. In this work, we choose Transformers as our modelling basis.

\paragraph{Learning.} The goal of training the ranking model is to create a vector space such that similar pair of dialogue contexts and responses have higher relevance score than the dissimilar ones. 

We train the ranking model with the same response selection data set $\mathcal{D}$ using the in-batch negative objective \cite{DBLP:conf/emnlp/KarpukhinOMLWEC20}. For a sampled batch of training data $\{(c_k, r_k)\}_{k=1}^b$, where $b$ is the batch size, the sampled contexts and responses are separately encoded using Eq. \eqref{eq:similarity} as $E_c(C)\in\mathbb{R}^{b\times n}$ and $E_r(R)\in\mathbb{R}^{b\times n}$, where $n$ is the output size of encoder modules. Next, the score matrix $S$ is computed as $E_c(C)^T E_r(R)\in\mathbb{R}^{b\times b}$. The in-batch negative objective \cite{DBLP:conf/emnlp/KarpukhinOMLWEC20} is then defined as minimizing the negative log likelihood of positive responses
\begin{equation}
    \mathcal{L}_G = -\frac{1}{b}\sum_{i=1}^{b}\log\frac{\exp(S_{ii})}{\exp(S_{ii}) + \sum_{j\ne i}\exp(S_{ij})},
\end{equation}
where $S_{ij}=G(c_i, r_j)$.

In this work, we build the context and response encoder with a 3-layer Transformers and its output size is 256. For all considered datasets, we pre-train the ranking model with a batch size $b=128$ for $20,000$ steps. For optimization, we use the Adam optimizer \cite{DBLP:journals/corr/KingmaB14} with a learning rate of 2e-5. For more details, we refer the readers to the original paper \cite{DBLP:conf/emnlp/KarpukhinOMLWEC20}.

\section{Hyper-parameter Setup}
In the following, we provide details on the search space for the hyperparameters. For number of negative responses $m$ in Eq. (1), the search space is \{$1$, \underline{$5$}, $10$, $15$, $20$\}, where the underline indicates the number selected based on the model performance on the validation set. The search space for the $p_{cc}(0)$ in corpus-level pacing function $p_{cc}(t)$ is \{$0.1$, $0.2$,  \underline{$0.3$}, $0.4$, $0.5$, $0.6$, $0.7$, $0.8$, $0.9$, $1.0$\}. For the $k_T$ in instance-level pacing function $p_{ic}(t)$, the search space is \{$1$,  $2$, \underline{$3$}, $4$, $5$, $\log_{10}^{500k}$\}, where $500k$ is the size of the training set. 

Each matching model is optimized with Adam optimizer \cite{DBLP:journals/corr/KingmaB14} with a learning rate of $2e$-$5$ and a batch size of 128. The total training step is set as $40,000$. $T$ in the corpus-level pacing fucntion $p_{cc}(t)$ and the instance-level pacing function $p_{ic}(t)$ is set as the half of the total training steps (i.e., $T=20000$).
\end{document}


\setlength{\arrayrulewidth}{0.7pt}

\clearpage
\section*{Supplementary Material}
\section{Ranking Model Training}
Here we provide more details on how to train the neural ranking model $G(\cdot, \cdot)$ that serves as the scoring basis in the proposed HCL framework.

\paragraph{Modelling.} Given a dialogue context $c$ and a response $r$, their context-response relevance score is defined as 
\begin{align}
\label{eq:similarity}
\begin{split}
    G(c, r) &= E_c(c)^T E_r(r).
\end{split}
\end{align}
Note that, the context $c$ is a long sequence which is acquired by concatenating all utterances in the dialogue context. The $E_c(c)$ and $E_r(r)$ are the context and response encoder. The context encoder $E_c(\cdot)$ takes the token sequence $c=c_0,...,c_{|c|}$ and returns the context representation $E_c(c)$ by taking the output state corresponds to the last token $c_{|c|}$. The same operation is applied when computing the response representation $E_r(r)$. In practice, the choice of $E(\cdot)$ could be any sequence model, e.g., LSTM \cite{DBLP:journals/neco/HochreiterS97}, RNN \cite{DBLP:journals/cogsci/Elman90},  Transformers \cite{DBLP:conf/nips/VaswaniSPUJGKP17}, and BERT \cite{DBLP:conf/naacl/DevlinCLT19}. In this work, we choose Transformers as our modelling basis.

\paragraph{Learning.} The goal of training the ranking model is to create a vector space such that similar pair of dialogue contexts and responses have higher relevance score than the dissimilar ones. 

We train the ranking model with the same response selection data set $\mathcal{D}$ using the in-batch negative objective \cite{DBLP:conf/emnlp/KarpukhinOMLWEC20}. For a sampled batch of training data $\{(c_k, r_k)\}_{k=1}^b$, where $b$ is the batch size, the sampled contexts and responses are separately encoded using Eq. \eqref{eq:similarity} as $E_c(C)\in\mathbb{R}^{b\times n}$ and $E_r(R)\in\mathbb{R}^{b\times n}$, where $n$ is the output size of encoder modules. Next, the score matrix $S$ is computed as $E_c(C)^T E_r(R)\in\mathbb{R}^{b\times b}$. The in-batch negative objective \cite{DBLP:conf/emnlp/KarpukhinOMLWEC20} is then defined as minimizing the negative log likelihood of positive responses
\begin{equation}
\label{eq:ranker}
    \mathcal{L}_G = -\frac{1}{b}\sum_{i=1}^{b}\log\frac{\exp(S_{ii})}{\exp(S_{ii}) + \sum_{j\ne i}\exp(S_{ij})},
\end{equation}
where $S_{ij}=G(c_i, r_j)$.

In this work, we build the context and response encoder with a 3-layer Transformers and its output size is 256. For all considered datasets, we pre-train the ranking model with a batch size $b=128$ for $20,000$ steps. For optimization, we use the Adam optimizer \cite{DBLP:journals/corr/KingmaB14} with a learning rate of 2e-5. For more details, we refer the readers to the original paper \cite{DBLP:conf/emnlp/KarpukhinOMLWEC20}.


\section{Hardware Specification}
The results reported in this paper in general are highly stable and should be easily replicable with our provided code. In Table 1, we present details regarding the hardware specifications and hyper-parameters setups for this work.

\begin{table}[H]
	\setlength{\abovecaptionskip}{3pt}
	\renewcommand\arraystretch{1.1}
	\centering  

    \begin{center}
    \scalebox{0.8}{
	\begin{tabular}{lr}
		\toprule
        Hardware& Specification\\
        \midrule
        RAM&CORSAIR\textsuperscript{\textregistered} Vengeance RGB PRO (16GB) $\times 4$\\
        CPU&AMD\textsuperscript{\textregistered} Ryzen 9 3900x 12-core 24-thread\\
        GPU&NVIDIA\textsuperscript{\textregistered} Tesla V100 Volta (32GB) $\times 1$\\
		\bottomrule
	\end{tabular}}
	\caption{Hardware specification of the used machine.}
	\end{center}
	\label{tb:hardware}
\end{table}

\section{Hyper-parameter Setup}
In the following, we provide details on the search space for the hyperparameters. For number of negative responses $m$ in Eq. (1), the search space is \{$1$, \underline{$5$}, $10$, $15$, $20$\}, where the underline indicates the number selected based on the model performance on the validation set. The search space for the $p_{cc}(0)$ in corpus-level pacing function $p_{cc}(t)$ is \{$0.1$, $0.2$,  \underline{$0.3$}, $0.4$, $0.5$, $0.6$, $0.7$, $0.8$, $0.9$, $1.0$\}. For the $k_T$ in instance-level pacing function $p_{ic}(t)$, the search space is \{$1$,  $2$, \underline{$3$}, $4$, $5$, $\log_{10}^{500k}$\}, where $500k$ is the size of the training set. 

Each matching model is optimized with Adam optimizer \cite{DBLP:journals/corr/KingmaB14} with a learning rate of $2e$-$5$ and a batch size of 128. The total training step is set as $40,000$. $T$ in the corpus-level pacing fucntion $p_{cc}(t)$ and the instance-level pacing function $p_{ic}(t)$ is set as the half of the total training steps (i.e., $T=20000$).

\clearpage
\bibliographystyle{acl_natbib}
\bibliography{anthology,acl2021}